\documentclass[runningheads]{llncs}
\usepackage[T1]{fontenc}
\usepackage{graphicx}

\usepackage{amsmath}
\usepackage{amsfonts}
\usepackage{booktabs}
\usepackage{caption}
\usepackage{subcaption}
\usepackage{tabularx}
\usepackage{hyperref}

\usepackage{color}

\newcolumntype{Y}{>{\centering\arraybackslash}X}
\newcolumntype{A}{ >{$} r <{$} @{} >{${}} l <{$} }
\begin{document}
\title{Scalability of Message Encoding Techniques for Continuous Communication Learned with Multi-Agent Reinforcement Learning}
\titlerunning{Scalability of Message Encoding for Cont. Comm. Learned with MARL}
%
\author{Astrid Vanneste\inst{\dag}\orcidID{0000-0002-6742-6722}, 
        Thomas Somers\inst{\dag}, \\
        Simon Vanneste\inst{\dag}\orcidID{0000-0002-9664-9925}, 
        Kevin Mets\inst{\dag \dag}\orcidID{0000-0002-4812-4841}, \\
        Tom De Schepper\inst{\dag \dag}\orcidID{0000-0002-2969-3133},
        Siegfried Mercelis\inst{\dag}\orcidID{0000-0001-9355-6566},  
        Peter Hellinckx\inst{*}\orcidID{0000-0001-8029-4720}}
\authorrunning{Astrid Vanneste et al.} 

%
%

\institute{
    University of Antwerp - imec \\ 
    IDLab - \textsuperscript{\dag}Faculty of Applied Engineering, \textsuperscript{\dag \dag}Department of Computer Science
    Sint-Pietersvliet 7, 2000 Antwerp, Belgium \\
    \textsuperscript{*}University of Antwerp, Faculty of Applied Engineering \\
    \email{
        \{astrid.vanneste,
        simon.vanneste,
        kevin.mets,
        tom.deschepper,
        siegfried.mercelis,
        peter.hellinckx\}@uantwerpen.be
    }
} 
%
\maketitle              
\begin{abstract}
Many multi-agent systems require inter-agent communication to properly achieve their goal. By learning the communication protocol alongside the action protocol using multi-agent reinforcement learning techniques, the agents gain the flexibility to determine which information should be shared. However, when the number of agents increases we need to create an encoding of the information contained in these messages. In this paper, we investigate the effect of increasing the amount of information that should be contained in a message and increasing the number of agents. We evaluate these effects on two different message encoding methods, the mean message encoder and the attention message encoder. We perform our experiments on a matrix environment. Surprisingly, our results show that the mean message encoder consistently outperforms the attention message encoder. Therefore, we analyse the communication protocol used by the agents that use the mean message encoder and can conclude that the agents use a combination of an exponential and a logarithmic function in their communication policy to avoid the loss of important information after applying the mean message encoder.

\end{abstract}
%
%

\section{Introduction}
Communication is an essential part of what makes humans intelligent and productive. The same thing can be said for multi-agent systems. The potential of these systems can be immensely improved by allowing inter-agent communication. Communication allows agents to overcome partial observability as well as coordinate their behaviour. In recent years, research has been done to allow the agents to learn a communication protocol themselves, perfectly tailored to the goal and environment of the agents. With an increasing or varying number of other agents, the need arises to summarize the contents of these messages in a fixed size encoding to make sure the agents can deal with this large or varying number of incoming messages.
In this paper, we investigate two ways to encode the messages: a mean communication encoder and an encoder which uses self-attention. We compare them to each other and to a no communication baseline. We investigate two different aspects of the environment. First we analyse the effect of increasing the amount of information that should be contained in a message. Secondly, we look at the effect of increasing the number of agents in the environment which also increases the number of incoming messages.

The remainder of this paper is structured as follows. First, in Section \ref{sec:related_work}, we take a look at prior, related work about communication learning methods. Section \ref{sec:background} provides some background knowledge about reinforcement learning and self-attention. In Section \ref{sec:methods}, we present the methods that we use in this work. Next, we show the various experiments performed and compare the results in Section \ref{sec:experiments}. We further discuss our results in Section \ref{sec:discussion}. Finally, we form a conclusion and present some future work in Section \ref{sec:conclusion} and \ref{sec:future_work} respectively.


\section{Related Work}\label{sec:related_work}

The research into communication learning using multi-agent reinforcement learning was introduced by Foerster et al. \cite{foerster2016learning} and Sukhbaatar et al. \cite{sukhbaatar2016learning}. Foerster et al. \cite{foerster2016learning} presented two different communication learning methods that learn discrete communication called Reinforced Inter-Agent Learning (RIAL) and Differentiable Inter-Agent Learning (DIAL). Sukhbaatar et al. \cite{sukhbaatar2016learning} proposed CommNet, a method to learn continuous communication between agents. Following their work, different methods to achieve inter-agent communication have been explored. Jaques et al.\cite{jaques2019socialinfluence} encourage communication that results in a change in the action policy of the agents. In MACC \cite{vanneste2021learning} this is taken a step further by performing counterfactual reasoning about the outcome of alternative messages to evaluate the message that was sent. Similar to CommNet and DIAL, A3C3 \cite{simoes2020a3c3} uses backpropagation to learn a communication protocol. A3C3 adds a centralized critic that learns a stationary value function which helps to learn the action and communication policies.  Lin et al. \cite{lin2021learning} use autoencoders to learn an encoding of the observation which will then be communicated to the other agents. 

Many of these works assume a constant or small number of agents. However, in many cases this is not realistic. When we want to communicate with a larger or variable number of agents, we have to create an encoding of the incoming messages. In CommNet \cite{sukhbaatar2016learning}, they take the mean of all incoming messages to account for a variable number of incoming messages. This approach was also taken by Singh et al. \cite{singh2018} in IC3Net. However, different techniques were also explored. ATOC \cite{jiang2018} uses a bi-directional LSTM to encode the incoming messages.  TarMAC \cite{das2019} uses a variation of the attention mechanism where both the sender and the receiver generate a value, which is then combined to generate the attention score for the message. This allows the agents to place varying importance on the incoming messages. The work of Peng et al. \cite{peng2018} specifically focuses on the challenge of encoding incoming messages while retaining all the necessary information. They propose an approach which combines a bi-directional recurrent neural network (RNN) and the attention mechanism. In these works, a variety of different message encoders have been presented. However, the scalability of these approaches has not been evaluated. In our work, we aim to give insight in how the mean message encoding and attention message encoding techniques perform when we increase the complexity of the environment as well as the number of agents. This will result in an increase in the information that has to be included in the messages and an increase in the number of incoming messages. 


\section{Background}\label{sec:background}
In this section, we provide some background information for our research. First, we introduce the theoretical framework on which this research is built. Next, we give a detailed explanation of the attention mechanism. 

\subsection{Markov Decision Processes}
The methods proposed in this paper use the decentralized Markov Decision Process (dec-MDP) as proposed by Oliehoek et al. \cite{oliehoek2016concise}. In a dec-MDP, at timestep $t$ each agent $a \in A$ receives an observation $o_t^a$ of the global state $s_t$ of the environment. Each agent individually is not able to observe the entire state of the environment. However, when we combine the information in the observations of all of the agents, we are able to construct the full state of the environment. This is called joint observability. Based on their observation, each agent will determine an action $u_t^a$. These actions will result in a new state $s_{t+1}$ and a reward for each agent $r_t^a$.

\subsection{Self-Attention}
\label{sec:background_attention}
The attention mechanism was introduced by Bahdanau et al. \cite{bahdanau2014neural}. Prior to this,  the most used sequence-to-sequence (seq2seq) mechanisms were recurrent neural networks (RNN)\cite{rumelhart1986rnn}. The big benefit of the RNN is that it is, in theory, capable of looking infinitely far back in the sequence. However, in practice when the length of the sequence increases, the RNN struggles to remember the information from the start of the sequence. Long Short Term Memory (LSTM) \cite{hochreiter1997lstm} addresses these concerns by using a gated architecture. However, it still has limits in the length of the sequence. 
Another downside of RNN's, more specific to the context of our work, is that it deals in sequences with a specific order. In our case, the incoming messages do not have a specific order and therefore the RNN architecture will be less suitable. For our research, we require an architecture that is designed to deal with a set of inputs of an undefined size. The requirements for this type of architecture are defined by Zaheer et al.\cite{zaheer2017}. The most important requirement is that the operation is permutation invariant. This means that the output does not change when we change the order of the input elements. 

Self-attention addresses the issues of RNN's and is widely used in natural language processing. It serves as one of the building blocks for the transformer proposed by Vaswani et al. \cite{vaswani2017}. Self-attention is able to compare the inputs with each other and calculate which inputs influence each other. Inside the scaled dot-product attention module presented by Vaswani et al.\cite{vaswani2017}, calculations are done in a couple of steps. First, we derive a key, query and value for each of the inputs. Next, we calculate the attention score by taking the dot product of the query of an input and all the keys and dividing by the square root of the dimension of the keys $d_k$. We take the softmax of all the attention scores that belong to the same query and multiply them with the values corresponding with the used key. Finally, we calculate the sum of all the weighted values to produce the output. To compute the attention output for multiple queries simultaneously, we pack the queries, keys and values into matrices $Q$, $K$ and $V$ respectively. We can then calculate the output according to Equation \ref{eq:attention} \cite{vaswani2017}. In Section \ref{sec:methods}, we discuss how we can use self-attention in the context of message processing.
\begin{equation}
    Attention(Q, K, V) = softmax\left(\frac{Q K^T}{\sqrt{d_k}}\right) V
    \label{eq:attention}
\end{equation}


\section{Methods} \label{sec:methods}

\subsection{CommNet}

\begin{figure}[t]
    \centering
    \includegraphics[width=0.45\linewidth, trim=5 7 23 15, clip]{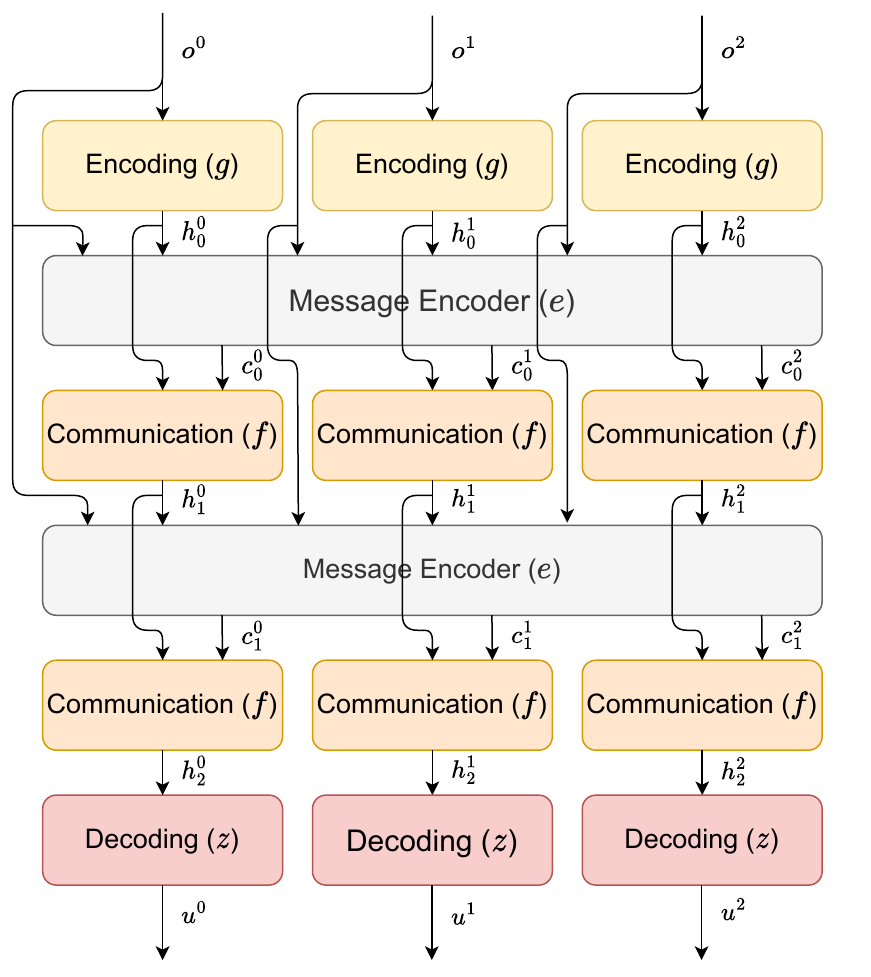}
    \caption{CommNet}
    \label{fig:commnet}
\end{figure}

CommNet \cite{sukhbaatar2016learning} is a communication learning method that allows the agents to send continuous messages to each other. Fig. \ref{fig:commnet} shows the architecture of CommNet when using two communication steps and three agents. We omit the time index since we only describe a single timestep in the environment. Subscripts are used to indicate the communication step index and superscripts are used to indicate the agent index. When the agent index is omitted, we describe the collection containing this variable for all agents. The architecture contains three different models, the encoding model, communication model and decoding model. First, the encoding model takes the observation and calculates the corresponding hidden state $h$. 
    \begin{equation}
        h^a_0 = g(o^a)
        \label{eq:encoder}
    \end{equation}
This hidden state is shared with the other agents. Since the hidden states are used as messages, we call the dimension of the hidden state the message size. All hidden states that arrive at the agent are encoded by the message encoder into a communication input $c$. The observation of the agent is passed to the encoding function because the observation often contains information that is needed to determine which information in the messages is relevant for the agent. It is important to note that, when calculating the message encoding for a certain agent, only the observation of that agent is used not the observations of the other agents.
\begin{align}
    \begin{split}
      c_i &= e(h_i, o)  \\&= [c_i^a: a \in A] \\&= [e^a(h_i, o^a): a \in A]  
      \label{eq:message_encoder}
    \end{split} 
    && (i \in \mathbb{N}: 0 \leq i < S)
\intertext{The communication network uses the hidden state and the communication input to calculate the next hidden state. This process can be repeated for $S$ communication steps.}
    h^a_{i+1} &= f(h^a_i, c^a_i) && (i \in \mathbb{N}: 0 \leq i < S)
    \label{eq:comm_step}
\end{align}
When all communication steps are completed the last hidden state is used by the decoding model to calculate a distribution over the action space. The action will be sampled from this distribution.
    \begin{equation}
        \pi^a = z(h^a_S)
        \label{eq:decoder}
    \end{equation}
During training, we first collect a batch of episodes using the current policy. This batch is then used to update our models using the loss function in Equation \ref{eq:loss_function}. The first part consists of the REINFORCE loss\cite{williams1992} where we reduce the variance on the reward by subtracting the mean of all rewards in the current batch as a baseline and dividing by the standard deviation of these rewards. The second part consists of an entropy loss that encourages the agent to explore, weighted by hyperparameter $\beta$. The communication channel is differentiable so the loss can be backpropagated through the communication channel to provide feedback to the other agents. 
\begin{equation}
    \mathcal{L} = -log(\pi(o^a | u^a))\left(\frac{r^a - \mu}{\sigma}\right) + \beta \sum_{u'^a}{\pi(o^a | u'^a) log(\pi(o^a | u'^a))}
    \label{eq:loss_function}
\end{equation}

\subsection{Message Encoding}
\label{sec:message_encoding}
In this paper, we investigate two different message encoding techniques namely the mean and self-attention. These techniques will define the behaviour of the message encoder described in Equation \ref{eq:message_encoder}.

\subsubsection{Mean Message Encoder}
The mean message encoder takes the hidden states of the other agents and takes the mean of them to calculate the output. Taking the mean of the hidden states is the approach originally proposed by Sukhbaatar et al. \cite{sukhbaatar2016learning}. The disadvantage of the mean encoder is that it gives each incoming message the same importance. Therefore, we cannot filter out irrelevant information or focus on specific information. Since the mean message encoder has less flexibility, we expect to see that information will be lost when creating the encoding. Therefore, given the mean of a set of messages it is not guaranteed that we can retrieve all the relevant information that was provided in these messages. For the mean message encoder, only the messages from the other agents are averaged. The message the agent sends will not be taken into account. To achieve this, we take the sum of all hidden states and add this to the vector containing the negative hidden states of the agents ($-h_i$). This results in a vector that, for each agent, contains the sum of the hidden states of the other agents. By dividing this by the number of other agents in the environment ($N-1$), we get a vector $c_i$ with, for each agent, the mean of the hidden states of the other agents. This can be seen in Equation \ref{eq:mean_encoder}.


\begin{equation}
    c_i = e(h_i, o) = \frac{1}{N-1} \left(- h_i + \sum_{a \in A}{h_i^a}\right)
    \label{eq:mean_encoder}
\end{equation}

\subsubsection{Attention Message Encoder}
The second message encoder is based on the attention mechanism. We use the scaled dot-product attention as proposed by Vaswani et al.\cite{vaswani2017} which was explained in Section \ref{sec:background_attention}. The key and value for the attention mechanism are calculated using the incoming messages. The query is calculated based on the message of the current agent and its observation. We do this because the observation may include information that determines which information contained in the received messages is relevant to the agent. 

\begin{equation}
    Q = q(h_i, o), K = k(h_i), V = v(h_i)
\end{equation}

\begin{equation}
    c_i = e(h_i, o) = Attention(Q, K, V)
\end{equation}

Since the attention mechanism is designed to be able to vary the importance of each of the inputs, the message encoder based on attention will be able to filter out unnecessary information. However, since the agents need to learn additional parameters to calculate the keys, queries and values, we expect training to be slower.


\section{Experiments}\label{sec:experiments}
In this section we describe our experiments. First, we explain the environment that was used in our experiments and how we scale this environment. Afterwards we analyse our results. In Figure \ref{fig:results_bar}, we can see a global overview of all the results. We first go into more detail for the results when we increase the number of labels and then we explain the results when increasing the number of agents. For each of the methods in all of the experiments we performed five runs with different random seeds. Our experiments are performed using RLlib \cite{liang2018rllib} and Tune \cite{liaw2018tune} which are built on the Ray framework \cite{moritz2017ray}. We use parameter sharing between agents because it has been shown to improve training performance \cite{foerster2016learning}.

\subsection{Matrix Environment} \label{Environment}
For the experiments, we use a matrix environment, inspired by the Matrix Communication Games presented by Lowe et al. \cite{lowe2019measuring}. The environment consists of $N$ agents and $L$ labels. At the start of an episode, two labels are selected from the pool of $L$ possible labels. Next, every agent randomly receives one of these two labels as its observation in a one-hot encoding. Because the distribution of the labels among the agents is random, it is possible that every agent receives the same label. The task of the agents is to say how many other agents received the same label. This results in a discrete action space. Each agent receives an individual reward of one if they correctly indicated the number of other agents with the same label or zero if they were unsuccessful. Therefore, the maximum reward that all of the agents can achieve together is equal to $N$. During our experiments we normalize this reward by dividing the total reward of the agents together by the number of agents. This way the maximum reward the agents can achieve together will always be equal to one. The episodes are only one timestep long so the agents have only one opportunity to find the correct answer.

This environment is jointly observable to the agents because they can only see their own observations and not the full state. However, all the observations of the agents combined gives the complete state of the environment. In order to succeed, the agents need to communicate their label with the other agents. The environment is easily scalable because we can increase the number of agents by increasing $N$ and increase the number of possible labels by increasing $L$. Increasing $N$ or $L$ will both have a different effect on the environment. By increasing the number of agents $N$, each agent will receive a larger number of messages and the action space will increase. When the number of labels $L$ increases, each agent will need to be able to communicate a larger number of different labels in their messages. 

\subsection{Baseline}
In addition to the two approaches with different message encoders, we also trained a group of agents that was not allowed to communicate. This clearly shows the importance of communication and whether or not the communicating agents are still benefiting from communication. The agents that cannot communicate will learn a policy that is nearly random. 
The no communication agents are implemented by using the same architecture as CommNet but without any communication steps. This means we only use the encoding and decoding networks without the communication network.

\begin{table}[t]
    \begin{minipage}[t]{.5\linewidth}
        \centering
        \caption{The hyperparameter values that are used in our experiments}
        \label{tab:hyperparameters}
        \begin{tabular}{|c|c|}
            \hline
            Hyperparameter                & Value   \\ \hline
            Learning Rate                 & 0.002   \\
            Discount Factor               & 0.99    \\
            Number of Communication Steps & 1       \\
            Train Batch Size              & 80      \\
            Activation                    & ReLU    \\
            Message Size                  & 16      \\ \hline 
        \end{tabular}
    \end{minipage}
    \begin{minipage}[t]{.5\linewidth}
        \centering
        \caption{Value of $\beta$ used to weigh the entropy loss for each experiment.}
        \label{tab:entropy_beta_values}
        \begin{tabular}{|c|c|}
            \hline
            Experiment          & $\beta$    \\ \hline
            $N = 3, L = 3$      & 0.44                  \\
            $N = 3, L = 8$      & 0.44                  \\
            $N = 3, L = 16$     & 0.44                  \\
            $N = 3, L = 24$     & 0.44                  \\
            $N = 8, L = 3$      & 0.15                  \\
            $N = 16, L = 3$     & 0.01                  \\
            $N = 24, L = 3$     & 0.01                  \\ \hline
        \end{tabular}
    \end{minipage}
\end{table}

\subsection{Hyperparameters and Network Architecture}

Our hyperparameters were determined empirically and using a grid search to get the best results for each experiment. The resulting hyperparameters are still mostly the same for each experiment and can be seen in Table \ref{tab:hyperparameters}. Table \ref{tab:entropy_beta_values} shows the weight $\beta$ for the entropy loss, which differs for each experiment. Throughout the experiments we kept the size of the networks and the message size constant. This was done to ensure a fair comparison and to clearly see the effect of scaling up the environment. The encoding model consists of a single linear layer with input size equal to the number of labels and output size equal to the message size, followed by a ReLU activation. The communication model is represented using a single linear layer with input size equal to twice the message size and output size equal to the message size, followed by a ReLU activation. Finally, the decoder model is a single linear layer with input size equal to the message size and the output size equal to the number of actions which is equal to the number of other agents, followed by a softmax function. The decoding model is the only model that can vary between experiments, since the action space changes when we change the number of agents, which will increase the output size of the network. The communication model will not be used in the experiments where communication is not allowed, making the complete network smaller. In addition to these models, the agents that use the attention message encoder will have three additional models to calculate the queries, keys and values. These three models are represented with a single linear layer and the output size is equal to the message size. The input size for the key and value networks is equal to the message size. For the query network the input size is equal to the message size plus the number of labels since the observation is included in the input.


\begin{figure}[t]
    \begin{subfigure}[t]{0.5\textwidth}
        \includegraphics[width=\linewidth]{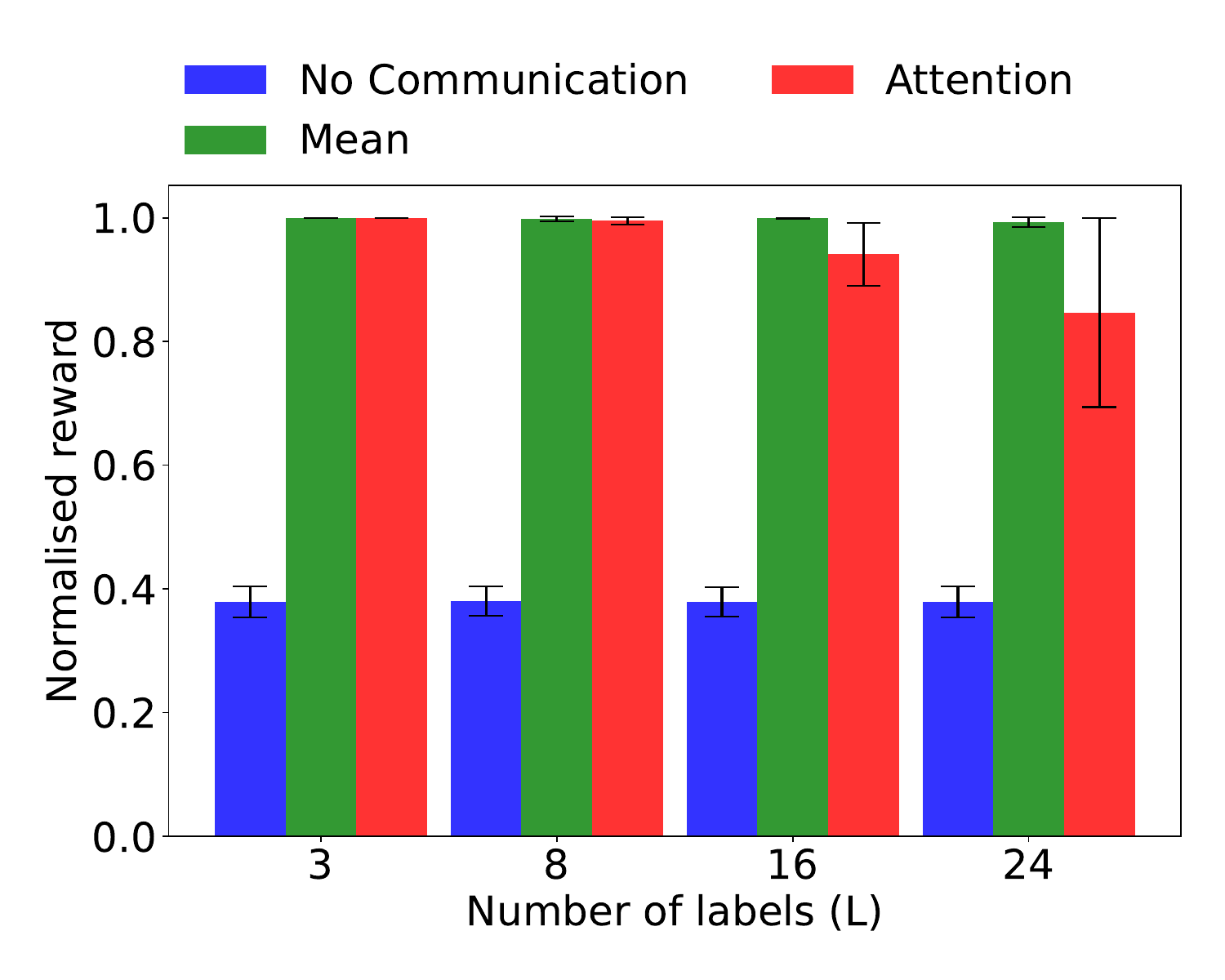}
        \caption{Scaling the number of labels with a constant number of agents ($N = 3$)}
        \label{fig:results_bar_increase_labels}
    \end{subfigure}
    \begin{subfigure}[t]{0.5\textwidth}
        \includegraphics[width=\linewidth]{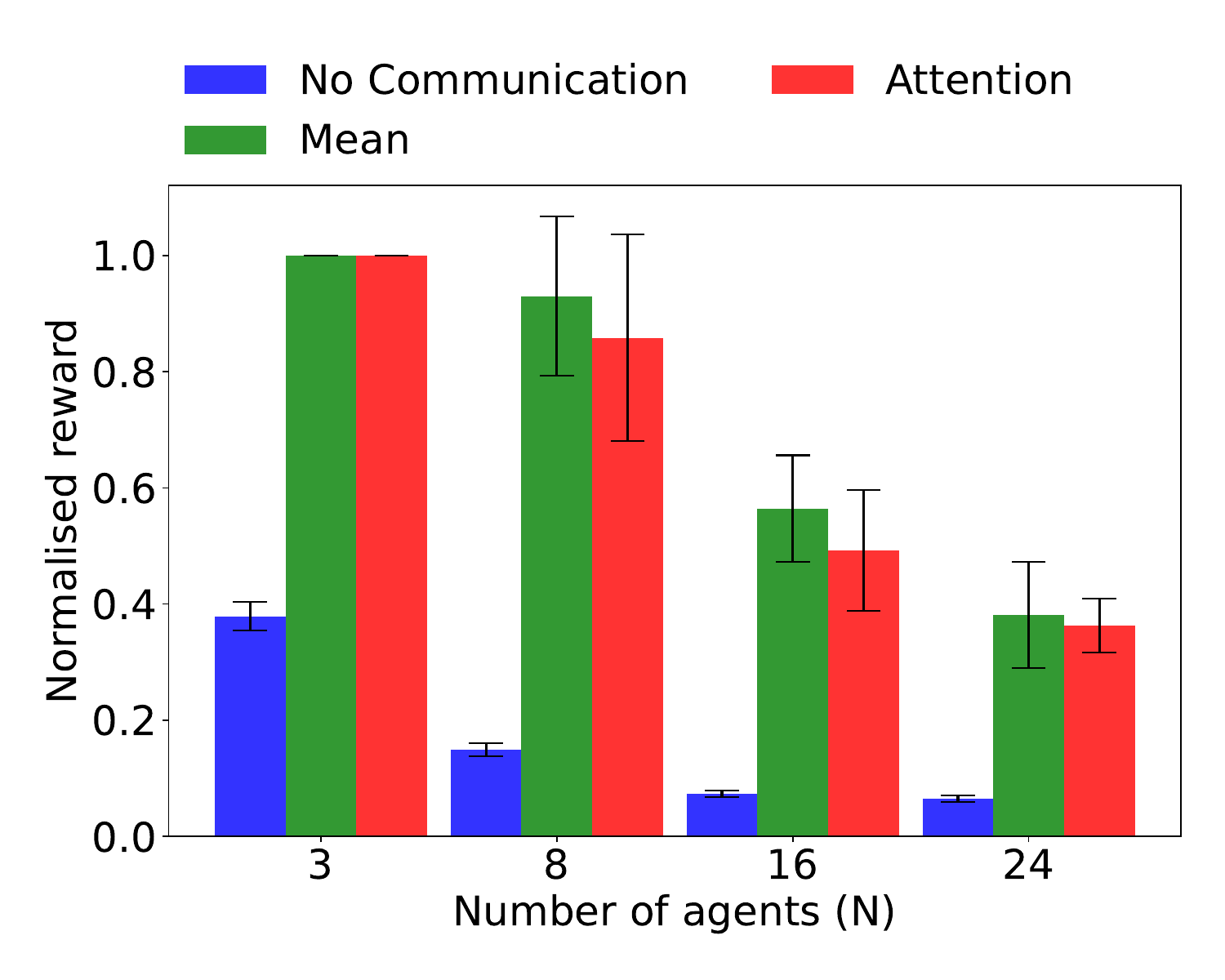}
        \caption{Scaling the number of agents with a constant number of labels ($L = 3$)}
        \label{fig:results_bar_increase_agents}
    \end{subfigure}
    
    \caption{Results in the matrix environment}
    \label{fig:results_bar}
\end{figure}

\subsection{Scaling the Number of Labels}
\begin{figure}[t]
    \begin{subfigure}[t]{0.5\textwidth}
        \centering
        \includegraphics[width=0.85\linewidth]{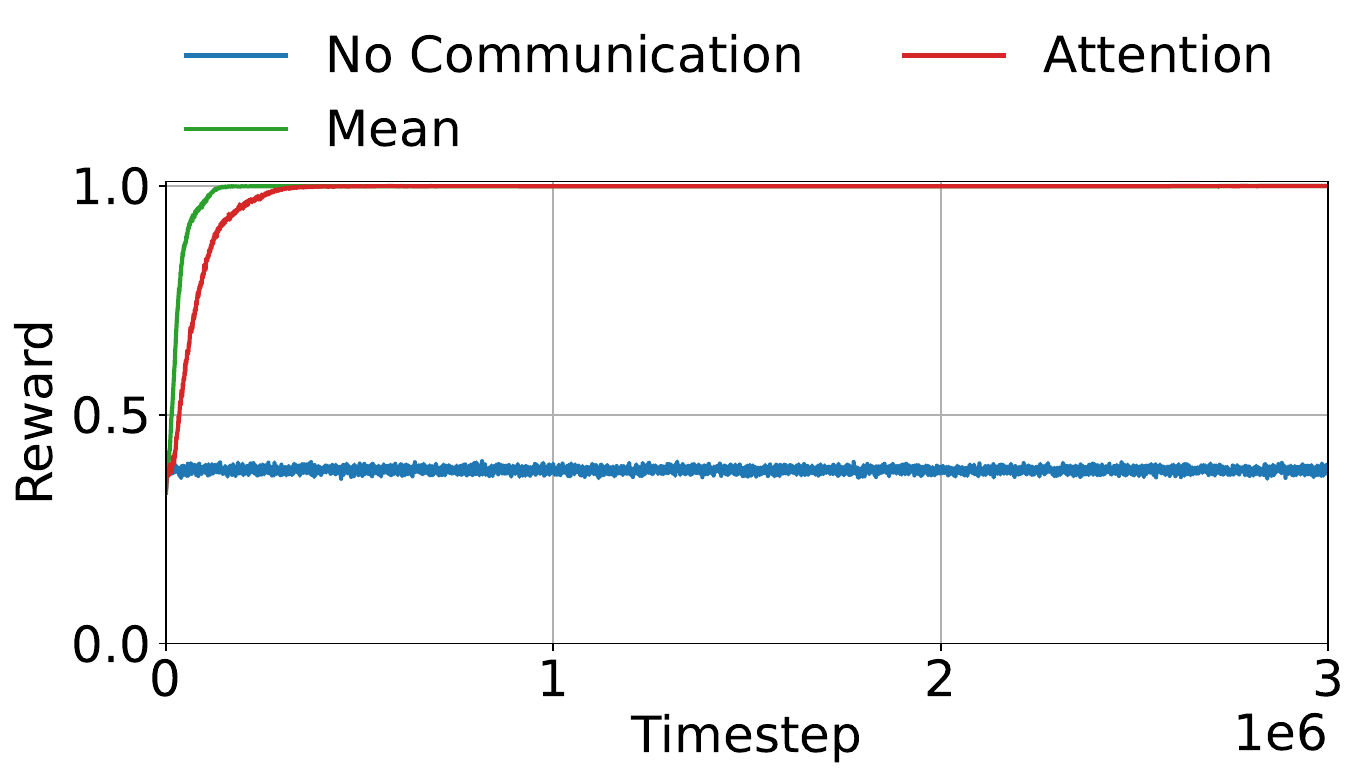}
        \caption{N = 3, L = 3}
        \label{fig:N3L3_scale_L}
    \end{subfigure}
    \begin{subfigure}[t]{0.5\textwidth}
        \centering
        \includegraphics[width=0.85\linewidth]{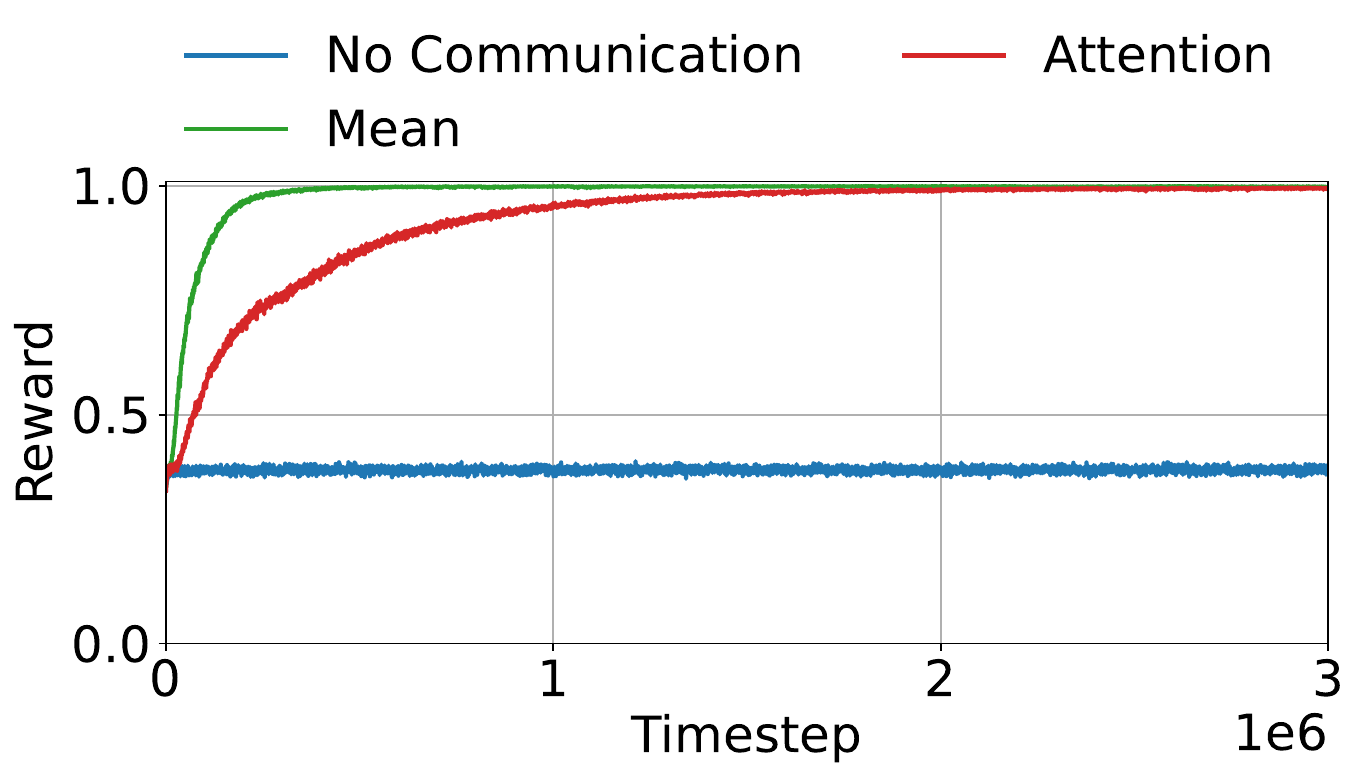}
        \caption{N = 3, L = 8}
        \label{fig:N3L8}
    \end{subfigure}
    \begin{subfigure}[t]{0.5\textwidth}
        \centering
        \includegraphics[width=0.85\linewidth]{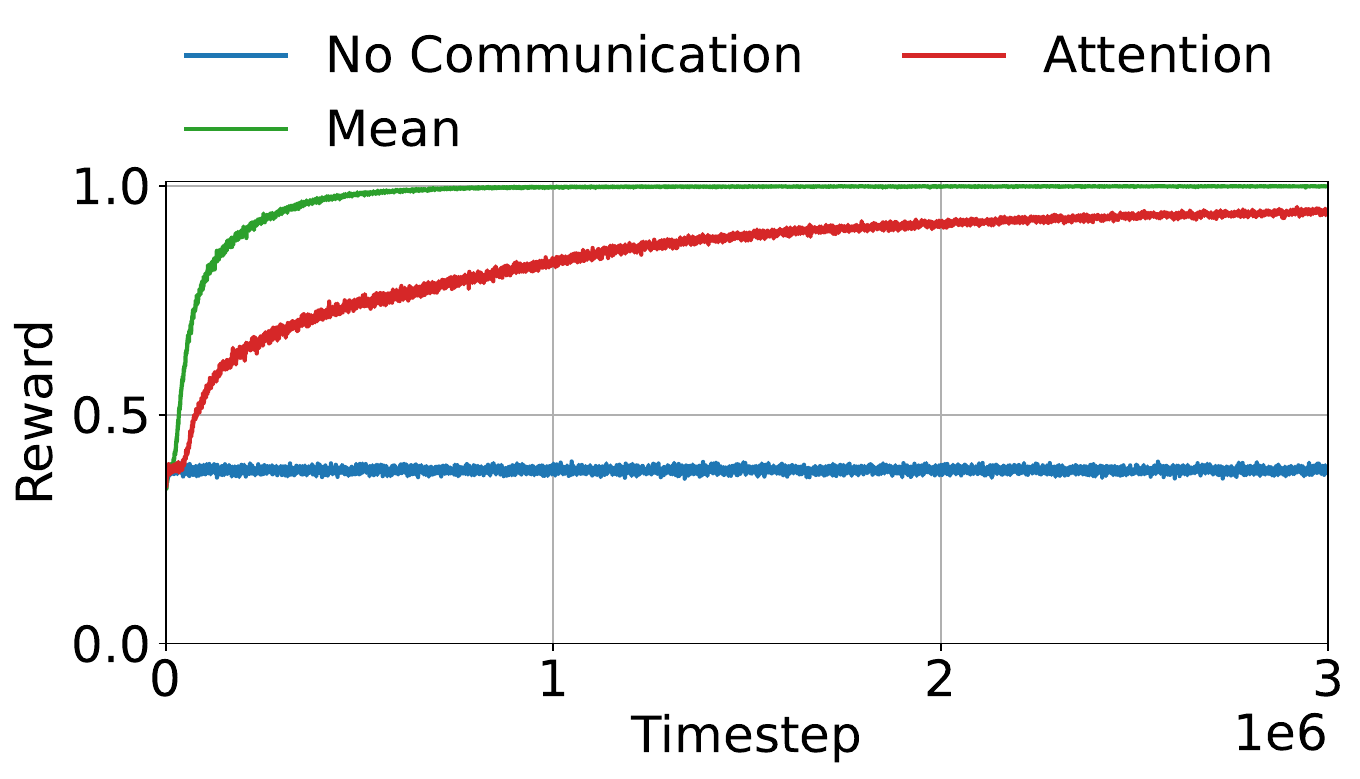}
        \caption{N = 3, L = 16}
        \label{fig:N3L16}
    \end{subfigure}
    \begin{subfigure}[t]{0.5\textwidth}
        \centering
        \includegraphics[width=0.85\linewidth]{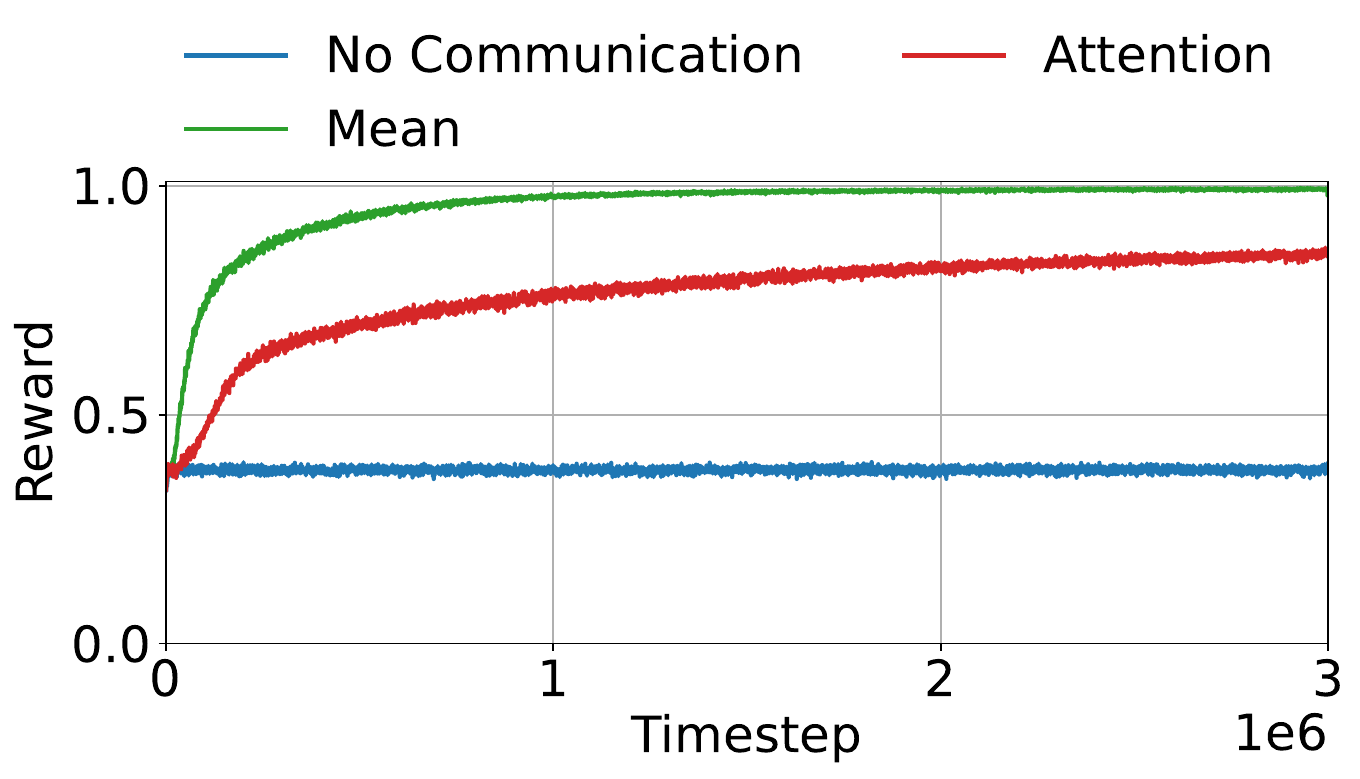}
        \caption{N = 3, L = 24}
        \label{fig:N3L24}
    \end{subfigure}
    
    \caption{Results when increasing the number of labels}
    \label{fig:results_increase_labels}
\end{figure}
In this section, we present a series of experiments where we increased the number of labels ($L \in \{3, 8, 16, 24\}$) while keeping the number of agents fixed ($N = 3$). We experimented with four different values for $L$ based on the message size ($= 16$), two values that are smaller than the message size ($L = 3$ and $L = 8$), one value that is equal to the message size ($L = 16$) and one value that is larger than the message size ($L = 24$). With a simple communication protocol that encodes the labels using a one-hot representation, we would need a message size equal to the value of $L$. Using this communication protocol both of the encodings should allow for an optimal action policy to be learned. For a number of labels that is higher than the message size, we would expect the mean message encoder to lose information when multiple message combinations result in the same mean message.

Figure \ref{fig:results_bar_increase_labels} shows the average performance and the standard deviation during the final 10\% of training when we increase the number of labels. The performance of the agents that do not communicate is slightly higher than what the performance would be of a random agent. This is because the agents learn that some scenarios are more common than others and gain a small benefit from that.  We see that for low values of $L$ ($L = 3$ and $L = 8$), both encodings are able to achieve the maximum reward. However, when the value of $L$ increases we see that the performance of the attention message encoder starts to diminish. The performance of the mean message encoder drops only 0.7\% between the experiment for $L = 3$ and the experiment for $L = 24$ while the performance of the attention based message encoder drops 12.6\%. 
In the setup of our experiments, we expected the performance of the mean message encoder to drop when the number of labels becomes higher than the message size due to the fact that every message has the same importance. If the mean message encoder can no longer preserve all relevant information, the attention message encoder should outperform the mean message encoder since it is a lot more flexible and can selectively change the attention values to change the importance of each message. However, in our results we can see that the mean message encoder always outperforms the attention message encoder even for 24 labels. However, we can see that once the number of labels becomes higher than the message size, the performance of the mean message encoder also starts to decrease.

Figure \ref{fig:results_increase_labels} shows the evolution of the reward during training for each configuration. We can see that across all of the experiments, the attention message encoder trains slower than the mean message encoder. Both the mean and attention message encoders get slower when we increase the value of $L$. However, this effect is more prevalent in the results of the attention message encoder.  

\subsection{Scaling the Number of Agents}

\begin{figure}[t]
    \begin{subfigure}[t]{0.5\textwidth}
        \centering
        \includegraphics[width=0.85\linewidth]{img/graphs/N3L3.pdf}
        \caption{N = 3, L = 3}
        \label{fig:N3L3_scale_N}
    \end{subfigure}
    \begin{subfigure}[t]{0.5\textwidth}
        \centering
        \includegraphics[width=0.85\linewidth]{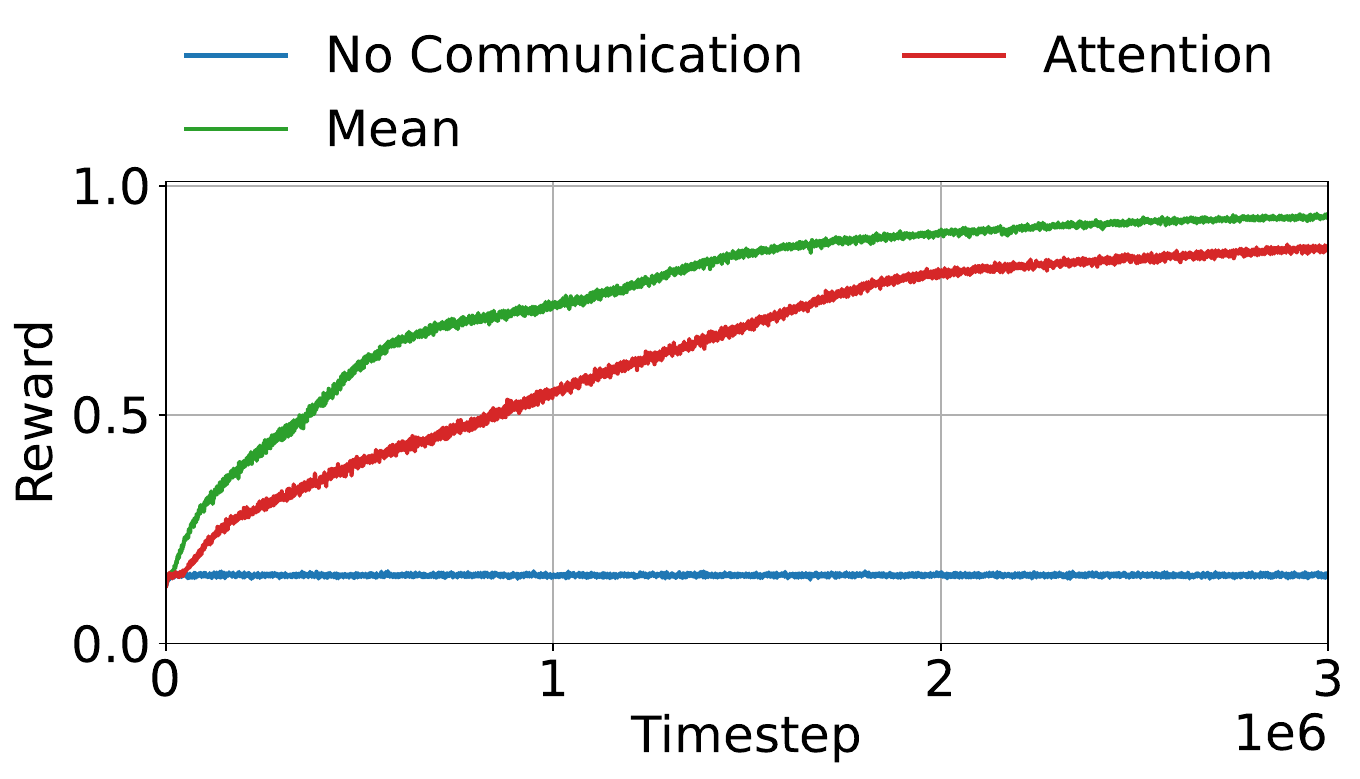}
        \caption{N = 8, L = 3}
        \label{fig:N8L3}
    \end{subfigure}
    \begin{subfigure}[t]{0.5\textwidth}
        \centering
        \includegraphics[width=0.85\linewidth]{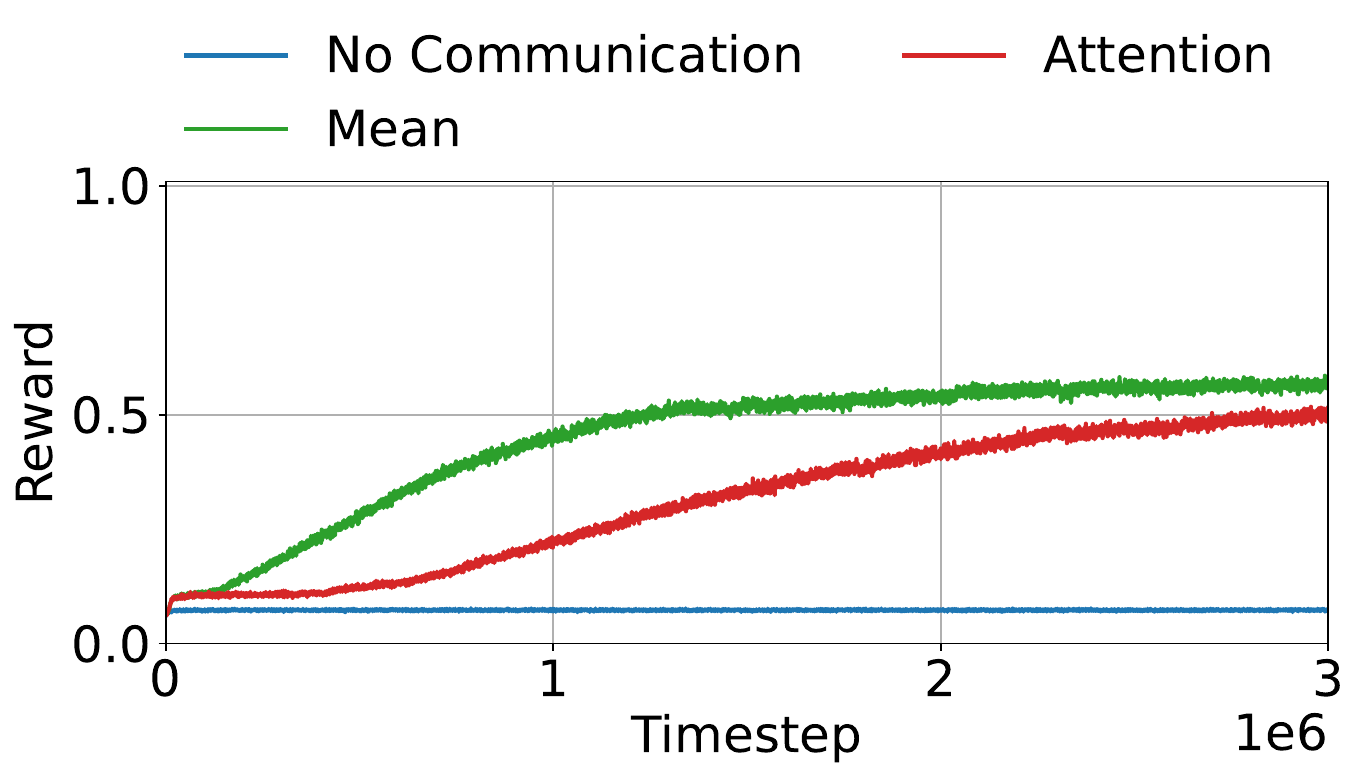}
        \caption{N = 16, L = 3}
        \label{fig:N16L3}
    \end{subfigure}
    \begin{subfigure}[t]{0.5\textwidth}
        \centering
        \includegraphics[width=0.85\linewidth]{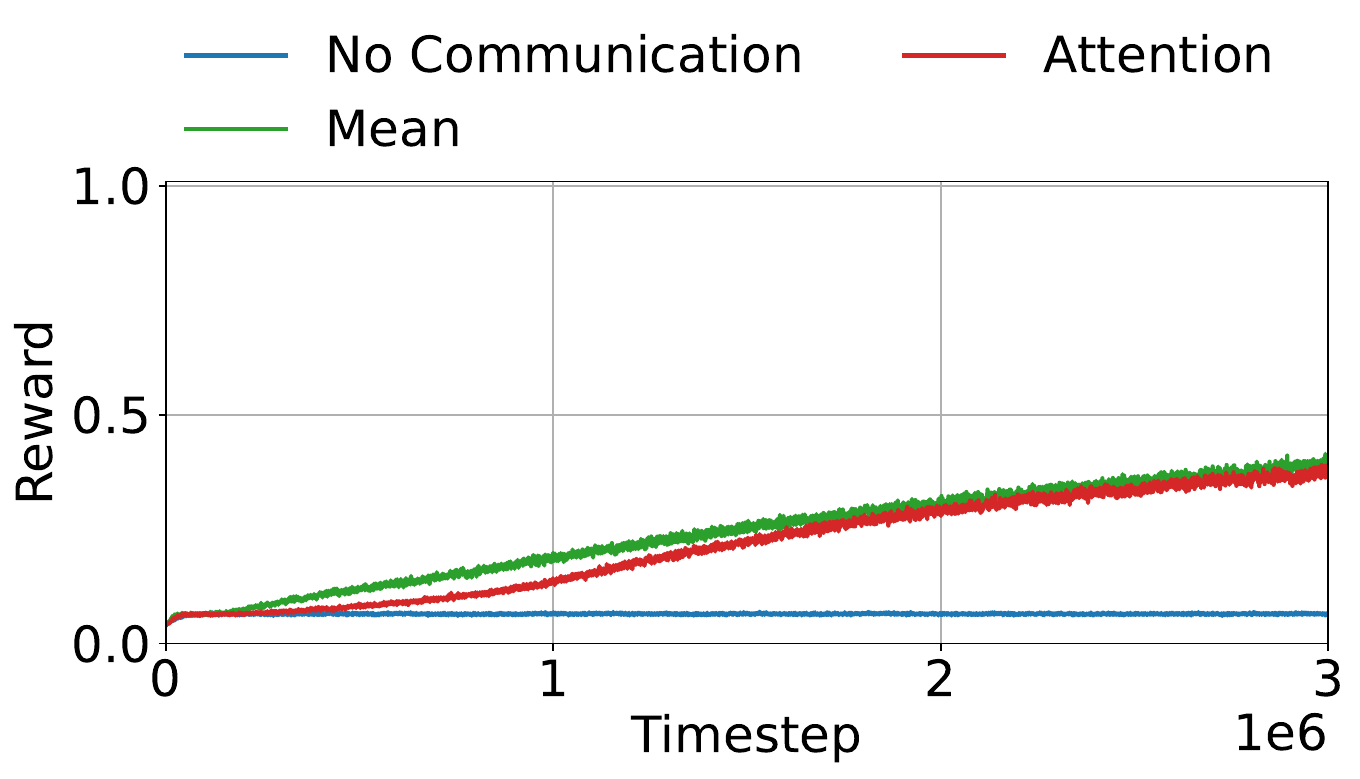}
        \caption{N = 24, L = 3}
        \label{fig:N24L3}
    \end{subfigure}
    
    \caption{Results when increasing the number of agents}
    \label{fig:results_increase_number_agents}
\end{figure}
In this section, we present a series of experiments where we increased the number of agents ($N \in \{3, 8, 16, 24\}$) while keeping the number of labels fixed ($L = 3$). The message size is the same as in the previous experiments ($=16$).
Figure \ref{fig:results_bar_increase_agents} shows the average performance and the standard deviation during the final 10\% of training when we increase the number of agents. Figure \ref{fig:results_increase_number_agents} shows the evolution of the reward during training for each configuration. In Figure \ref{fig:results_bar_increase_agents}, we see that when we increase the number of agents all of the methods perform worse. With an increasing number of agents, the action space becomes larger as well. This makes the problem a lot more complex. The agents that are not allowed to communicate still have a near random policy which results in a decreasing performance when the action space becomes larger. When we compare the reward of the mean message encoder and the attention message encoder we can see a clear difference. The drop in reward is significantly larger for the attention message encoder. Even though the attention message encoder still has the lowest performance for 24 agents, we see that the difference becomes significantly smaller. In Figure \ref{fig:results_increase_number_agents} we see that the attention message encoder trains slower than the mean message encoder. Again, we see that this difference becomes larger at first but decreases again for 24 agents. 

\begin{table}[t]
\centering
\caption{Communication policy for the matrix environment with $N = 3$, $L = 8$ and a message size of four}
\label{tab:comm_policy}
\begin{tabularx}{\linewidth}{|Y|YYYY|}
\hline
Label       & Message[0]       & Message[1]       & Message[2]       & Message[3]       \\ \hline
0           & 2,0319           & 0,0000           & 1,9082           & 0,0000           \\
1           & 2,1820           & 0,0000           & 5,2146           & 0,0000           \\
2           & 12,2357          & 0,0000           & 7,6871           & 0,0000           \\
3           & 5,7066           & 0,0000           & 6,5708           & 0,0000           \\
4           & 3,3961           & 0,0000           & 5,5810           & 0,0000           \\
5           & 7,7053           & 0,0000           & 6,9608           & 0,0000           \\
6           & 1,9294           & 0,0000           & 3,7184           & 0,0000           \\
7           & 2,1637           & 0,0000           & 0,0000           & 0,0000           \\ 
\hline
\end{tabularx}
\end{table}

\begin{figure}[t]
    \begin{subfigure}[t]{0.5\linewidth}
        \centering
        \includegraphics[width=0.85\linewidth, trim=27 0 45 15, clip]{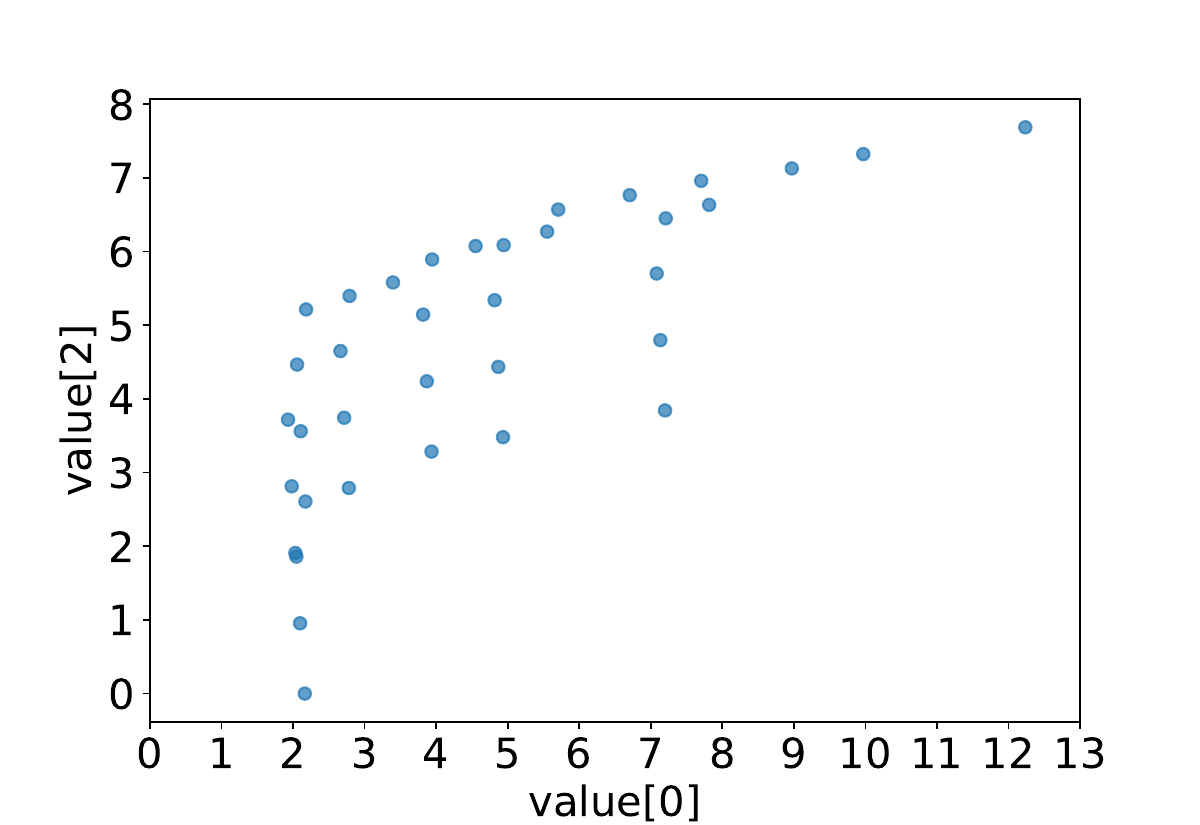}
        \caption{The mean values for each possible combination of two messages}
        \label{fig:scatter_means_individual}
    \end{subfigure}
    \begin{subfigure}[t]{0.5\linewidth}
        \centering
        \includegraphics[width=0.85\linewidth, trim=27 0 45 15, clip]{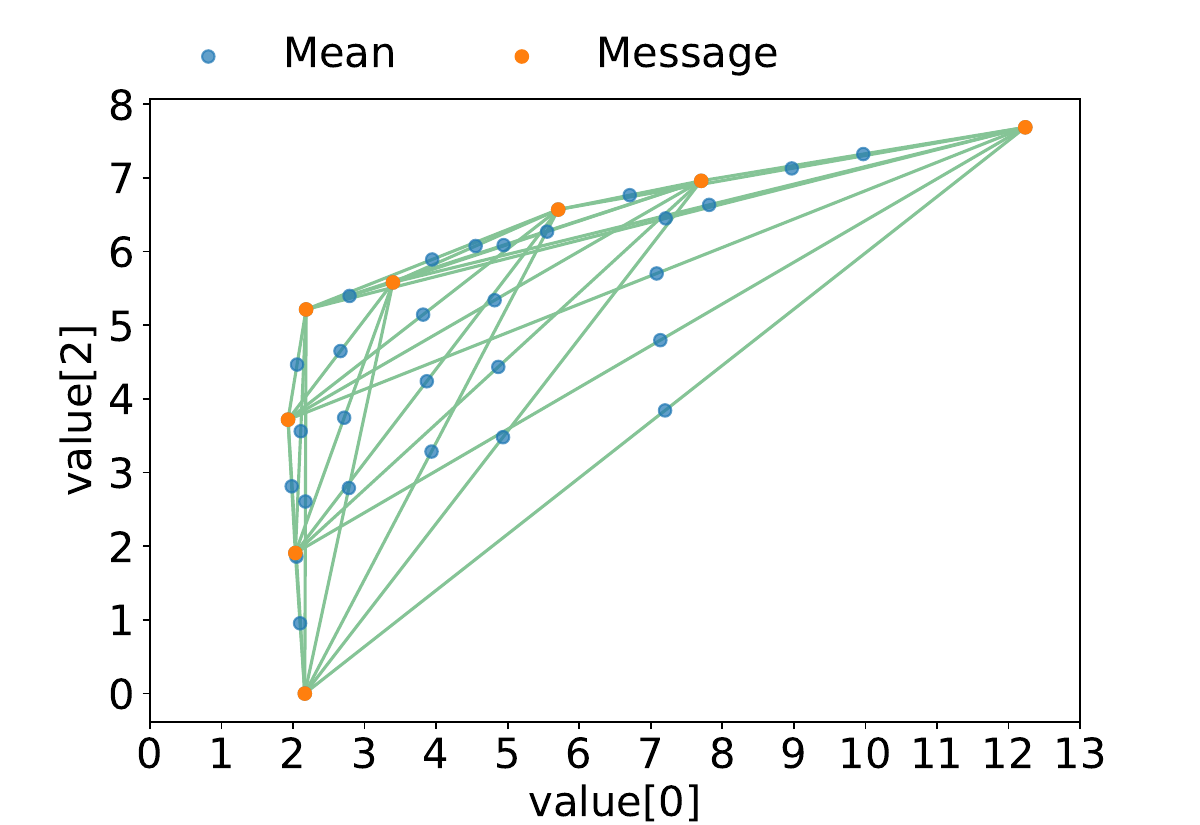}
        \caption{The mean values for each possible combination of two messages connected to the message values that produced the mean value.}
        \label{fig:scatter_means_interconnections}
    \end{subfigure}
    \caption{Messages and the mean value of the possible combinations of two messages}
    \label{fig:scatter_means}
\end{figure}

\begin{figure}[t]
    \begin{subfigure}[t]{0.5\linewidth}
        \centering
        \includegraphics[width=0.8\linewidth, trim=7 7 7 15, clip]{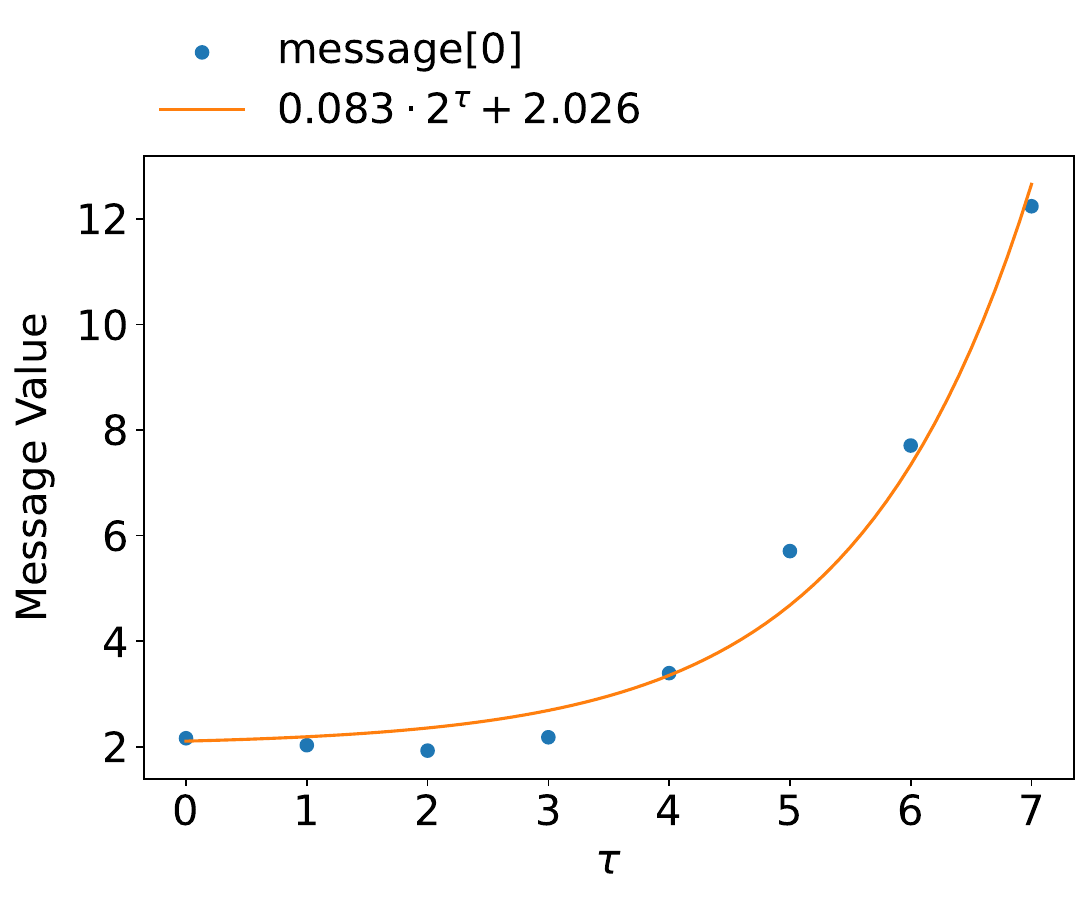}
        \caption{message[0]}
        \label{fig:regression_message_0}
    \end{subfigure}
    \begin{subfigure}[t]{0.5\linewidth}
        \centering
        \includegraphics[width=0.8\linewidth, trim=7 7 7 15, clip]{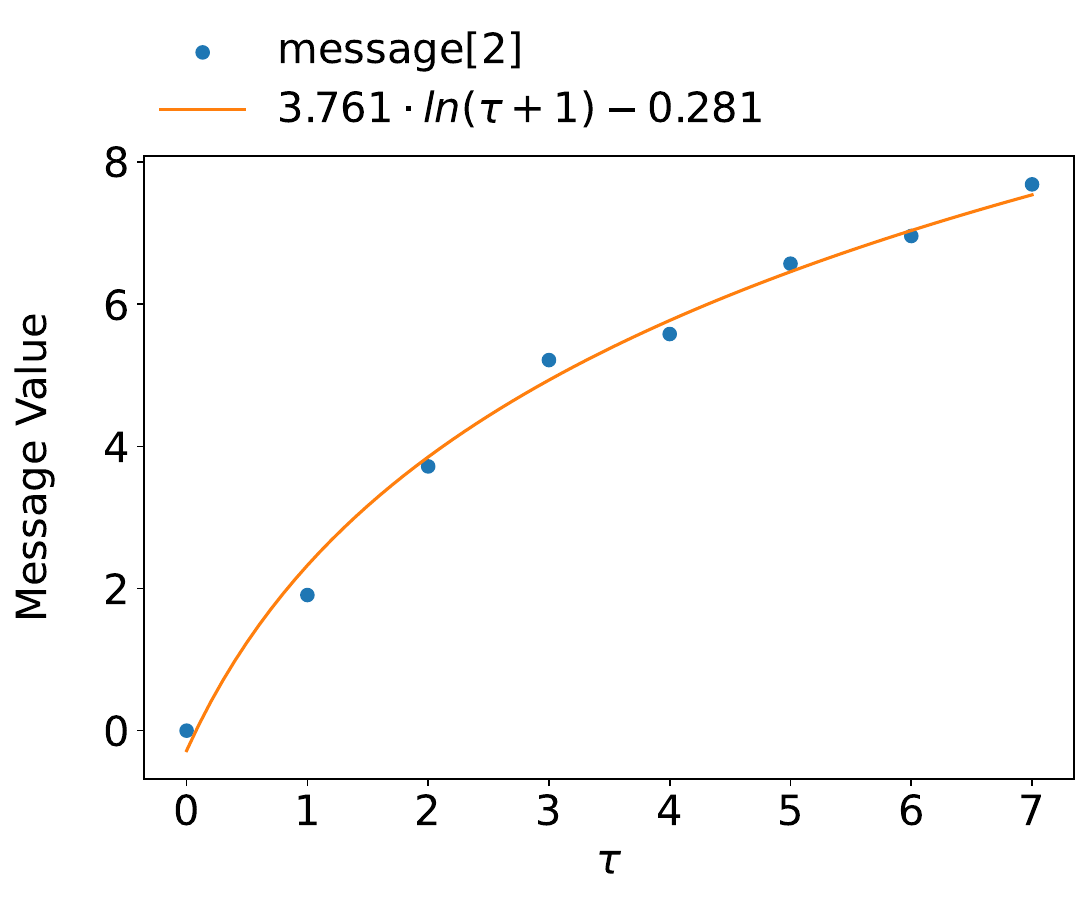}
        \caption{message[2]}
        \label{fig:regression_message_2}
    \end{subfigure}
    \caption{The actual message values in function of parameter $\tau$ compared to the function that best approximates the data according to our regression results.}
    \label{fig:regression_comm_policy}
\end{figure}

\subsection{Communication Analysis}
To gain further insight in our results, we analyse the communication policy that is learned using the mean message encoder. For this purpose we use a simple version of the matrix environment with $N = 3$, $L = 8$ and a message size of four. Since the message size is smaller than $L$, the agents cannot fall back on a one hot encoding of the labels. Table \ref{tab:comm_policy} shows the communication policy that the agents have learned when they consistently achieve the goal. We see that the agents only require two of the available four numbers to communicate the label info. Using this communication policy we can calculate all the possible mean values for all possible message combinations. These values can be seen in Figure \ref{fig:scatter_means}. Here we see that each of the means is separable from the other values with the exception of two message combinations that cause a very similar mean value. The distance between these two mean points is $5.11 \times 10^{-2}$. This shows that the agents will be able to reliably determine the global state of the environment, except in this specific case.

If we want to describe the curve of the message values in Figure \ref{fig:scatter_means_interconnections}, we can use a parametric equation where we describe $x$ and $y$ in function of a common parameter $\tau$. We choose $\tau$ in such a way that it increases as we go along the curve. We can plot both message values in function of parameter $\tau$. This can be seen in Figure \ref{fig:regression_comm_policy}. The curves resemble an exponential function for message[0] and a logarithmic function for message[2]. Therefore, we can write the parametric equation of our message values as displayed in Equation \ref{eq:parametric}. 

\begin{equation}
    \begin{cases}
        x = a \cdot  2^\tau + b \\
        y = c  \cdot ln(\tau + 1) + d
    \end{cases}
    \label{eq:parametric}
\end{equation}

We perform least squares linear regression to determine values for $a$, $b$, $c$ and $d$ that will result in the curves that most closely match the message data. This results in the following equation:
\begin{equation}
    \begin{cases}
        x = 0.083 \cdot  2^\tau + 2.026 \\
        y = 3.761 \cdot ln(\tau + 1) - 0.281
    \end{cases}
    \label{eq:parametric_after_regression}
\end{equation}

These regressions match our message data very well ($R^2 = 0.98$ for the regression of message[0] and $R^2 = 0.99$ for the regression of message[2]). Figure \ref{fig:regression_comm_policy} shows the message values in function of $\tau$ alongside the functions described in Equation \ref{eq:parametric_after_regression}.
In Figure \ref{fig:scatter_means_interconnections}, we can see that the communication policy is very well suited to result in mean values, positioned at the midpoint of the line connecting the two messages, that can be separated from each other. By representing this curve as a parametric equation, we have determined that each of the message values can be represented using a different non-linear function. By learning a representation combining an exponential function and a logarithmic function for each of the message values, the agents were able to retain all necessary information in the encoding that the mean message encoder generates. 

\section{Discussion}
\label{sec:discussion}

In our experiments, we evaluated the performance of agents with either a mean message encoder or an attention message encoder. Across all the experiments, we saw that the mean message encoder outperforms the attention message encoder. Even when the number of labels exceeds the message size, the agents were able to find a communication protocol that achieves very good performance. By analysing the communication protocol in a small scale experiment,  we were able to determine that the agents choose to represent the labels using a combination of an exponential and a logarithmic function. This ensures that no relevant information is lost. 

We performed two types of experiments. In the first we increased the number of labels that the agents need to be able to communicate while in the second series of experiments we increased the number of agents. In the results summarized in Table \ref{tab:results}, we see that the mean message encoder is not affected much by the increase in the number of labels while the performance of the attention message encoder clearly suffers. The difference in performance between the mean message encoder and the attention message encoder keeps growing as we increase the number of labels. However, when looking at the second series of experiments, we see that the mean message encoder is affected by increasing the number of agents since the action space will grow as well. We can also see that the difference in performance between the mean message encoder and the attention message encoder does not keep growing as we increase the number of agents. 

\begin{table}[t]
\centering
\caption{Results for the different evaluated scenarios. We show the reward normalized according to the number of agents to get the reward per agent. We also show the percentage change in average reward when going from the mean message encoder to the attention message encoder.}
\label{tab:results}
\begin{tabularx}{\linewidth}{|Y|YYY|Y|}
\hline
                  & No Comm.                 & Mean                   & Attention                   & $\Delta$                       \\ \hline
$N = 3$, $L = 3$  & $0.379 \pm 0.025$        & $1.000 \pm 0.000$      & $1.000 \pm 0.000$           & $0.000\%$                      \\
$N = 3$, $L = 8$  & $0.380 \pm 0.024$        & $0.998 \pm 0.004$      & $0.995 \pm 0.006$           & $-0.301\%$                     \\
$N = 3$, $L = 16$ & $0.379 \pm 0.024$        & $0.999 \pm 0.001$      & $0.941 \pm 0.051$           & $-5.806\%$                     \\
$N = 3$, $L = 24$ & $0.379 \pm 0.025$        & $0.993 \pm 0.008$      & $0.847 \pm 0.153$           & $-14.703\%$                    \\ \hline
$N = 3$, $L = 3$  & $0.379 \pm 0.025$        & $1.000 \pm 0.000$      & $1.000 \pm 0.000$           & $0.000\%$                      \\
$N = 8$, $L = 3$  & $0.149 \pm 0.011$        & $0.930 \pm 0.137$      & $0.858 \pm 0.178$           & $-7.742\%$                     \\
$N = 16$, $L = 3$ & $0.073 \pm 0.006$        & $0.564 \pm 0.092$      & $0.492 \pm 0.104$           & $-12.766\%$                    \\
$N = 24$, $L = 3$ & $0.065 \pm 0.006$        & $0.381 \pm 0.091$      & $0.363 \pm 0.046$           & $-4.724\%$                     \\ 
\hline
\end{tabularx}
\end{table}


\section{Conclusion}\label{sec:conclusion}
In this work, we evaluated the difference in performance between a mean message encoder and an attention message encoder when we increase the complexity of the environment and the number of agents. Intuitively, the attention approach seems the most ideal in this area since it can vary the importance of the different incoming messages. However, in the evaluated scenarios of the proposed matrix environment, we see that the mean message encoder consistently outperforms the attention message encoder. We were able to analyse the communication protocol of the agents that use the mean message encoder in a small scale scenario. The results showed that the agents use an exponential and a logarithmic function to avoid the loss of important information. 


\section{Future work}\label{sec:future_work}
This work provides initial results that compare different message encoding techniques. For a full understanding of the advantages and disadvantages of each technique, we need to look into some more aspects of the message encoding problem. In this paper, we focus on the comparison between the mean technique and the attention technique. However, there are more possibilities. RNN's and a combination of RNN's and attention have successfully been applied in the past \cite{jiang2018}\cite{peng2018}. So far, we have only looked into continuous message encoding. The encoding of discrete messages poses some additional challenges. Due to the limited number of available messages, the risk of information loss is a lot higher for discrete communication. Therefore, in future work we want to look at more message encoding techniques and apply these to both continuous and discrete messages. Additionally, we also want to investigate their performance using different communication learning techniques. Finally, the environment used in our work has a constant number of agents and therefore the agents will receive a constant number of incoming messages at every timestep. Episodes in the matrix environment are also only one timstep long. In the future, we want to use more complex environments with a varying number of agents and longer episodes. 


\section*{Acknowledgements}
Astrid Vanneste and Simon Vanneste are supported by the Research Foundation Flanders (FWO) under Grant Number 1S12121N and Grant Number 1S94120N respectively.

\typeout{} 
\bibliographystyle{splncs04}
\bibliography{references}
\end{document}